\documentclass[10pt,twocolumn,letterpaper]{article}

\usepackage{cvpr}              

\usepackage[dvipsnames]{xcolor}

\usepackage{amsmath}
\usepackage{multirow}
\usepackage{graphicx}
\usepackage[symbol]{footmisc}

\newcommand{\method}{\texttt{DynOMo}\@\xspace}

\newcommand{\PAR}[1]{\noindent{\bf #1~}}

\usepackage{pifont}
\newcommand{\cmark}{\ding{51}}%
\newcommand{\xmark}{\ding{55}}%

\definecolor{cvprblue}{rgb}{0.21,0.49,0.74}
\usepackage[pagebackref,breaklinks,colorlinks,citecolor=cvprblue]{hyperref}

\def\paperID{153} 
\def\confName{3DV\xspace}
\def\confYear{2025\xspace}

\title{
DynOMo: Online Point Tracking by Dynamic Online Monocular Gaussian Reconstruction
\vspace{-0.3cm}
}

\author{Jenny Seidenschwarz$^{1,2,3}\footnotemark{}$
\quad
Qunjie Zhou$^{4}$
\quad
Bardienus P. Duisterhof$^{3}$
\quad
Deva Ramanan$^{3}$
\quad
Laura Leal-Taix\'{e}$^{4}$\\
\\
$^1$\text{Technical University of Munich}
\hspace{0.5cm}
$^2$\text{MCML}
\hspace{0.5cm}
$^3$\text{Carnegie Mellon University}
\hspace{0.5cm}
$^4$\text{NVIDIA}
}

\begin{document}

\twocolumn[{%
\renewcommand\twocolumn[1][]{#1}%
\maketitle
\begin{center}
\vspace{-0.6cm}

\includegraphics[width=0.95\linewidth]{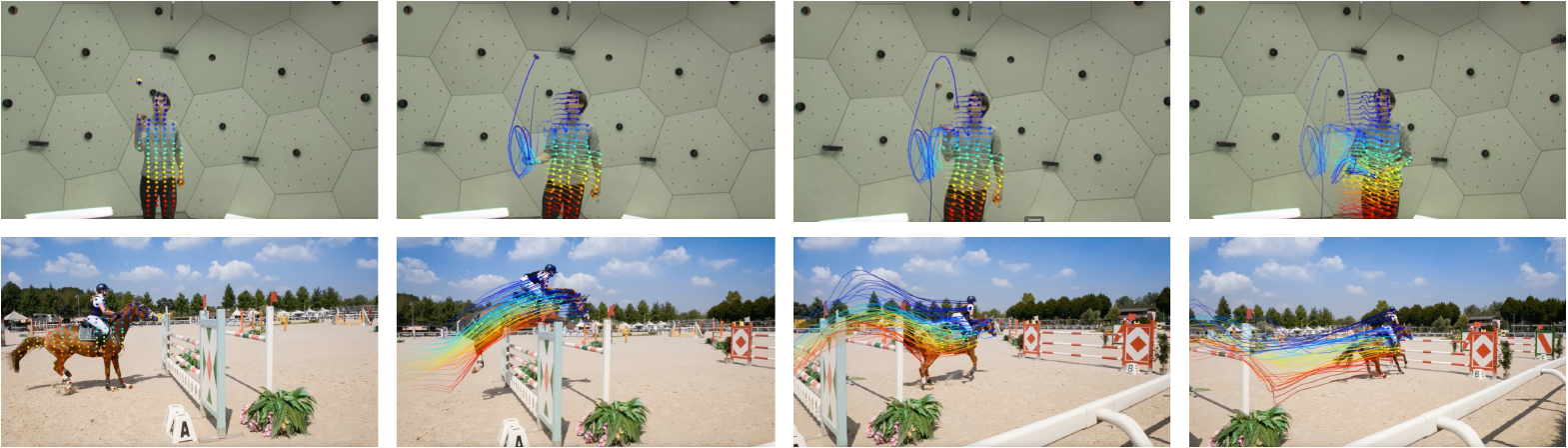}
\vspace{-0.2cm}
\captionof{figure}{
\textbf{Monocular Online Point Tracking}:
In this work, we present \method for the task of monocular online point tracking from \textit{pose-free} videos through joint 3D reconstruction and camera localization based on a dynamic 3D Gaussian representation. Please find the code and more visualizations on our project page {\footnotesize\url{https://jennyseidenschwarz.github.io/DynOMo.github.io}.}
}
\vspace{-0.1cm}
\end{center}%
}]
%

\begin{abstract}
\vspace{-0.4cm}
Reconstructing scenes and tracking motion are two sides of the same coin. Tracking points allow for geometric reconstruction \cite{keetha2024splatam}, while geometric reconstruction of (dynamic) scenes allows for 3D tracking of points over time \cite{luiten2023dynamic,som2024shapeofmotion}. The latter was recently also exploited for 2D point tracking to overcome occlusion ambiguities by lifting tracking directly into 3D \cite{wang2023omnimotion}. 
However, above approaches either require offline processing or multi-view camera setups both unrealistic for real-world applications like robot navigation or mixed reality. 
We target the challenge of \textbf{online 2D and 3D point tracking} from unposed monocular camera input introducing \textbf{Dy}namic \textbf{O}nline \textbf{Mo}nocular Reconstruction (\textbf{\method}). We leverage 3D Gaussian splatting to reconstruct dynamic scenes in an online fashion. Our approach extends 3D Gaussians to capture new content and object motions while estimating camera movements from a single RGB frame. \textbf{\method} stands out by enabling emergence of point trajectories through robust image feature reconstruction and a novel similarity-enhanced regularization term, without requiring any correspondence-level supervision. It sets the first baseline for online point tracking with monocular unposed cameras, achieving performance on par with existing methods. We aim to inspire the community to advance online point tracking and reconstruction, expanding the applicability to diverse real-world scenarios. 
\footnotetext[1]{Correspondence to \texttt{j.seidenschwarz@tum.de}}

\end{abstract}

\vspace{-1cm}
\section{Introduction}
\label{sec:intro}
Modeling static geometry and dynamic object motion over a long-term time range is highly correlated to point tracking.
In fact, they can even be addressed by point tracking and vice versa. 
Early point tracking methods have been primarily utilized in Simultaneous Localization and Mapping (SLAM)~\cite{durrant2006slam, davison2007monoslam, campos2021orb3} to simultaneously reconstruct a 3D scene map and localize a camera moving through the scene. SLAM approaches typically focus on static scene contents to ensure accurate localization.
To eliminate the influence of dynamic objects, they often track sparse features that lie on static elements in the scene. Recently, works like \cite{keetha2024splatam, Matsuki2024gsslam, Li2023denseslam, Zhu2022niceslam, zhu2024nicerslam} also explored dense SLAM for static scenes based on 3D Gaussians Splatting~\cite{kerbl3Dgaussians} (3DGS) and Neural Radiance Fields~\cite{mildenhall2020nerf}. 

{\bf 2D Point Tracking.} To model dynamic content, TAP-Vid \cite{doersch2022tapvid} recently introduced the challenge of ``Tracking Any Point" in 2D across frames of a video.
Building upon prior works~\cite{sand2008particle}, recent approaches \cite{harley2022particle,doersch2023tapir, doersch2022tapvid} learn correspondence matchers that can be augmented to exploit correlation between multiple tracks with a transformer~\cite{karaev2023cotracker}.
%
%
However, such methods require training data with correspondence-level annotations. Because of the difficulty in labeling, such motion annotations are often limited to pure 2D pixel correspondences. 
Yet motion projected in 2D is inherently more ambiguous due to the discarded depth information \cite{som2024shapeofmotion}. 
{\bf 3D Point Tracking.} As such, much recent work proposes to track points directly in a (pseudo) 3D space~\cite{wang2023omnimotion,luiten2023dynamic,som2024shapeofmotion} by removing the need for training data and instead optimizing a spacetime reconstruction directly on the test video of interest. 
Omni-motion~\cite{wang2023omnimotion} and the concurrent work of shape-of-motion~\cite{som2024shapeofmotion}
optimize a dynamic reconstruction of an entire scene in an offline fashion, requiring 2D flow and 2D point tracks
as supervision, respectively. Such offline 2D supervision and batch optimization can be compute intensive and prone to 2D errors, as described above. Conversely, D-3DGS\cite{luiten2023dynamic} achieves \textit{online} point tracking by formulating the optimization as a dynamic 3D Gaussian reconstruction. However, D-3DGS requires a large number (27) of static cameras and an initial 3D point cloud of the scene. Thus existing approaches are difficult to apply to real-world scenarios like robot navigation and mixed reality, which require online and interactive tracking of both static and dynamic content.
Towards that end, we present \method, the first online point tracking algorithm for monocular videos in-the-wild (with unknown camera poses). 
Our approach draws inspiration from SpaTAM~\cite{keetha2024splatam}, D-3DGS~\cite{luiten2023dynamic}, and 3D feature distillation~\cite{kobayashi2022distilledfeaturefields}, combining the strengths of dynamic scene modeling, online tracking, and camera localization using Gaussian Splatting.
%
Our key insight is to augment the RGB reconstruction loss with pre-trained 2D pixel encoders, including monocular depth, semantic object class labels, and feature descriptors. Interestingly, we make use of {\em no} temporal supervision, and demonstrate that 3D tracking {\em emerges} naturally from static 2D encoders by simply augmenting each dynamic 3D Gaussian with a descriptor (that is inferred during online optimization).
%
Crucially, descriptors can be used to {\em group} Gaussians when applying spatial regularizers during the optimization. Past work often groups Gaussians by proximity~\cite{luiten2023dynamic}, but we find it better to group by descriptor (i.e., points should move like other points with similar appearance rather than those are simply nearby). Our experiments corroborate the well-known result from perceptual grouping that proximity is the result, not the cause, of similarity \cite{fowlkes2003affinity}. 
%
Because our optimization explicitly recovers camera pose and new regions of the scene via densification~\cite{keetha2024splatam}, our online reconstructions allow for exploration of newly emerging scene content while keeping track of previously observed regions. 

In summary, our model can track any point in a dynamic scene over time through 3D reconstruction from monocular video input. Our main contributions are as follows:
%
(i) We introduce a \textit{tracking-by-reconstruction} baseline for tackling online TAP (tracking any point) in 2D and 3D, with unposed monocular cameras.
(ii) We demonstrate that temporal correspondence can emerge without correspondence-level optical flow supervision or known camera poses in challenging monocular settings.
(iii) We show through evaluation on common benchmarks that our approach significantly outperforms baseline methods in the same online monocular setting and is competitive with state-of-the-art offline 2D trackers.
%
We see our work as an initial step towards online TAP and reconstruction with unposed monocular cameras in the wild, and we encourage future research to further enhance its accuracy and efficiency.
\section{Related Work}
\label{sec:rel_work}
\subsection{Dynamic Scene Reconstruction}
\vspace{-0.1cm}

\label{subsec:rw_dynamic_scene}
\PAR{Neural Radiance Fields.} NeRF~\cite{mildenhall2020nerf} has given rise to a new class of scene representations for novel view synthesis. While its earlier work was limited to static scenes~\cite{martin2021nerfw, barron2022mipnerf360, muller2022instantnerf, rematas2022urf}, more recent works ~\cite{gao2021dynnerf, pumarola2021dnerf, li2021neuralsceneflowfields,Li22023dynibar,Park201nerfies,Park2021hypernerf,Xian2021spacetime, yang2023emernerf} have extended the topic to also model dynamic scenes. However, NeRF represents a scene implicitly and thus requires time-consuming volumetric rendering.

\PAR{3D Gaussian Splatting.} 3DGS~\cite{kerbl3Dgaussians} has recently gained great attention in neural rendering. In contrast to NeRF, it explicitly represents a scene with 3D Gaussian primitives and uses efficient differentiable rasterization techniques. Following a similar road map, 3DGS methods started from handling only static scenes but recently expanded to modeling dynamic scenes\cite{kratimenos2024dynmf,li2023spacetime,wu20234dgaussians,yang2023deformable3dgs,yang2023gs4d,duisterhof2023MDSplat, luiten2023dynamic, sun20243dgstream}. In contrast to NeRFs, the fast per-sequence optimization of 3DGS allows for more real-world use cases. 
Hence, our work adopts 3DGS to achieve dense reconstruction in dynamic scenes in a monocular setting towards real-world applications. 

\subsection{Point Tracking}
\vspace{-0.1cm}

\PAR{Simultaneous Localization and Mapping.}  Establishing point tracks over a video sequence is the core task of SLAM~\cite{durrant2006slam, davison2007monoslam}. However, exisiting SLAM algorithms focus on reconstructing static scenes for accurate geometry and camera motion estimation. Sparse SLAM~\cite{durrant2006slam, davison2007monoslam, campos2021orb3} commonly relies on keypoint detection and matching to tracking points between frames. Recent works leverage dense scene representations such as NeRF~\cite{Sucar2021imap, Li2023denseslam, kong2023vmap, Zhu2022niceslam, zhu2024nicerslam} and 3DGS~\cite{Matsuki2024gsslam, keetha2024splatam} to perform dense SLAM via pose optimization from view synthesize on static scenes.
Our method builds on top of SplaTAM~\cite{keetha2024splatam} to enable its reconstruction on dynamic scenes. 


\PAR{Tracking Any Point.} 
In contrast to the SLAM task, the task of tracking any point aims to track both static and dynamic points in a scene.
The seminal work Partical Video~\cite{sand2008particle} models video motion using a set of particles that move through time where they refine optical flow estimates for long-term consistency with occlusion handling. Its recent revisit PIPs~\cite{harley2022particle} upgrades the particle motion estimates with a feed-forward network in an offline sliding-window manner. TAPIR~\cite{doersch2023tapir} combines PIPs with a better matching model inspired by TAPNet~\cite{doersch2022tapvid}, while PointOdyssey~\cite{zheng2023PointOdyssey} provides an updated version of PIPs.
Instead of tracking each point independently~\cite{doersch2022tapvid,harley2022particle,doersch2023tapir}, CoTracker~\cite{karaev2023cotracker} exploits the correlation between all image tracks with a transformer network, resulting in significantly improved accuracy. 

Different from the above works that operate tracking in the 2D image space, several recent work~\cite{Xiao2024spatialtracker, wang2023omnimotion, luiten2023dynamic} propose to lift 2D points into 3D and perform tracking in the 3D space to better handle occlusions and maintain temporal and spatial coherence. 
Omnimotion \cite{wang2023omnimotion} optimizes a volumetric representation per video with view synthesize and optical flow supervisions at test time. 
SpaTracker~\cite{Xiao2024spatialtracker} leverages an off-the-shelf monocular depth estimator to lift pixels into 3D with associated CNN features and construct a triplane representation for dense feature sampling. Then the authors train a transformer to update the 3D trajectory over local windows of temporal frames based on their triplane features in a feed-forward manner.
D-3DGS~\cite{luiten2023dynamic} extends static 3D Gaussian splatting~\cite{kerbl3Dgaussians} to cope with dynamic scenes. With the explicit Gaussian representation, reconstruction on both static and dynamic scene points over time naturally solves the point tracking task in a global 3D space. Different from ~\cite{Xiao2024spatialtracker, wang2023omnimotion}, it is supervised using view synthesize and physics-inspired regularization without the need of correspondence-level supervisions. Another recent work~\cite{som2024shapeofmotion} refines long-term 2D tracks from \cite{doersch2023tapir} by lifting them to 3D Gaussians using depth information meanwhile enable novel view synthesize.

\PAR{Dynamic Online Monocular Point Tracking.}
Most point tracking methods are designed for offline tracking except for SLAM methods and D-3DGS~\cite{luiten2023dynamic}. However, SLAM methods struggle with dynamic scenes 
and D-3DGS~\cite{luiten2023dynamic} requires posed videos in a multi-camera setting. 
Compared to them, our goal is to perform online point tracking with unposed monocular videos, which enables point tracking for real-time response applications such as live video analysis, surveillance systems and mixed reality applications.



\section{Method}
\label{sec:method}

\begin{figure*}[htp!]
    \centering
    \includegraphics[width=\linewidth]{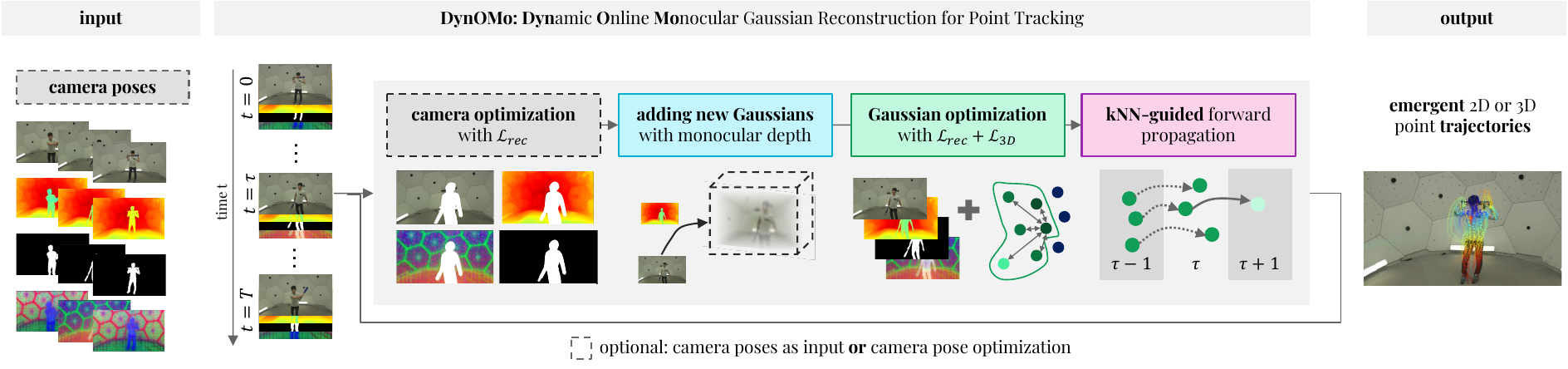}
    \caption{\textbf{Tracking points with online dynamic monocular reconstruction:} Our pipeline assumes an input video sequence, (predicted) depth maps, sparse segmentation masks as well as image features as input. In our online reconstruction pipeline we optimize for the camera pose $\mathcal{C}$, add a set of new Gaussians based on the densification concept \cite{keetha2024splatam}, optimize all Gaussians together and forward propagate $\mathcal{G}$ and $\mathcal{C}$. Finally, we directly extract 3D point trajectories from single Gaussians $G_p$ and project them to the image plane to obtain 2D trajectories.
    }
    \label{fig:pipeline_overview}
\end{figure*}


In this section, we present \textbf{\method}, our online point tracking pipeline via dynamic 3D Gaussian reconstruction for unposed monocular camera videos. 
Building on top of \cite{luiten2023dynamic, keetha2024splatam}, we combine the powerful 3DGS-based dynamic scene representation and the flexible online tracking paradigm for pose-free videos through simultaneous scene reconstruction and camera localization.
The key to \method{'s} performance lies in three technical adaptions for the online tracking setting, namely, 1) reconstruction signal enhancement with stronger image features and depth supervisions, 2) semantic-based foreground and background separation to enable camera tracking, and 3) motion regularization bootstrapping via a feature-similarity-guided weighting mechanism. 

In the following we first detail how we represent the scene via 3DGS in \cref{subsec:method_scene_repr} and explain the online tracking pipeline in \cref{subsec:method_pipeline}, followed by details about optimization supervisions (\cf \cref{subsec:rec_loss} and \cref{subsec:motion_regularizer}).
Finally, we show how to estimate 2D and 3D trajectories from a query point in \cref{subsec:trajectory_est}.



\subsection{Dynamic Gaussian Scene Representation} 
\vspace{-0.1cm}

\label{subsec:method_scene_repr}
We first introduce how we model a dynamic scene using 3D Gaussians. We start with a recap on 3D Gaussian splatting (3DGS) technique on static scene and its recent extension for dynamic scene representaiton by D-3DGS~\cite{luiten2023dynamic}. 
Finally, we present our adaption of D-3DGS to cope with challenges faced with monocular online point tracking.

\PAR{Preliminaries: 3DGS for static scenes.} 
3D Gaussian Splatting~\cite{kerbl3Dgaussians} represents the appearance and geometry of 
a static 3D scene with a set of $M$ explicit anisotropic 3D Gaussian distributions $\mathcal{G} = \{G_i\}_{i=0}^M$. Each 3D Gaussian is parametrized by its mean $\mu_i$ and covariance matrix $\Sigma_i$ and defined as:

\begin{equation}
    G_i(x; \mu_i, \Sigma_i) = e^{-\frac{1}{2}(x-\mu_i)^T\Sigma_i^{-1}(x-\mu_i)}.
\end{equation}
The covariance matrix is defined as $\Sigma_i = R_is_is_i^TR_i^T$ where $s_i$ is  a scale vector and $R_i$ a rotation matrix represented by a quaternion vector $q_i$. 
Each $G_i$ is additionally assigned with color $c_i$ and opacity $o_i$ attributes. 
%
%
To efficiently render a RGB image at a given camera, the 3D Gaussians are splatted into 2D Gaussians $G_{i, 2D}(p;\mu_{i, 2D}, \Sigma_{i, 2D})$ on the image plane utilizing an affine approximation of the camera projection \cite{Zwicker2001VolumeSplatting}. 
Finally, the color of a pixel $p$ is computed by $\alpha$-composing the color values of the projected 2D Gaussians 
that intersect with pixel $p$:
\begin{equation}
    C_p = \sum_{i \in H} T_i \alpha_i c_i
\end{equation}
where $\alpha_i = o_i e^{-\frac{1}{2}(p-\mu_{i,2D})^T\Sigma_{i,2D}^{-1}(p-\mu_{i,2D})}$ 
and $T_i = \prod_{j=1}^{i-1} (1-\alpha_j)$. 
The splatting-based rendering process is fully differentiable thus the Gaussian parameters $(\mu_i, q_i, s_i, c_i, o_i)$ are usually optimized directly by an image reconstruction loss~\cite{kerbl3Dgaussians} between a training image and the synthesized image rendered at that training camera pose.


\PAR{Preliminaries: 3DGS for Dynamic Scenes.} 
As mentioned in \cref{subsec:rw_dynamic_scene}, many recent works~\cite{kratimenos2024dynmf,li2023spacetime,wu20234dgaussians,yang2023deformable3dgs,yang2023gs4d,duisterhof2023MDSplat, luiten2023dynamic} explored how to model dynamic scenes with 3DGS. 
Among those, we adopt the formulation of D-3DGS~\cite{luiten2023dynamic} where a dynamic Gaussian at timestamp $\tau$ is defined by $G^\tau_i \equiv (\mu^\tau_i, q^\tau_i, s_i, c_i, o_i)$. Here $\mu^\tau_i$ and $q^\tau_i$ are the Gaussian mean and rotation at timestamp $\tau$ while the other parameters are constant over time as in the original 3DGS.
In contrast to its related work~\cite{yang2023deformable3dgs, li2023spacetime, yang2023gs4d} that require offline training of the scene representation over a whole video, D-3DGS is online optimized enabling online tracking-by-reconstruction via 3D Gaussians, which intersects with our interest.
However, this work requires an unrealistic multi-camera setting of 27 multi-view images per timestamp to guarantee quality reconstruction and tracking performance. 
Moving it directly to a monocular setting leads to significantly degraded performance as we show in \cref{tab:tracking_on_jono}.

\PAR{Adaptation for Monocular Online Reconstruction.}
In the monocular online reconstruction setting, each timestamp we only have access to  a monocular RGB image, which contains much sparser supervision signals to constrain the Gaussian optimization process compared to multi-view setting.
Therefore, to compensate this, we extend each Gaussian with two extra parameters, \ie, a semantic instance label $g_i \in R$ and a visual feature $f_i \in R^D$ which are constant overtime and optimized jointly with other parameters. Specifically, our monocular dynamic Gaussian is defined as $G^\tau_i \equiv (\mu^\tau_i, q^\tau_i, s_i, c^\tau_i, o_i, g_i, f^\tau_i)$ at a timestamp $\tau$. 
Inspired by recent correspondence-supervised 2D point tracking approaches \cite{doersch2023tapir,karaev2023cotracker,harley2022particle,Xiao2024spatialtracker}, we use the visual feature i) as a stronger version of photometric information to constrain the reconstruction to agree with each other in the feature space and ii) to provide weighting guidance for the optimization regularization.
We further explore semantic level information to assign an instance id for each Gaussian to decouple foreground and background particles. It not only allows us to enforce semantic-level consistency when a particle moves around but also supports us to design regularizations (\cf \cref{subsec:motion_regularizer}) specialized for the moving particles.

To initialize and optimize our augmented dynamic 3DGS parameters, in addition to the RGB frames, we assume to have access to extra information of that image specifically its image visual feature map extracted with DinoV2 \cite{Oquab2024dinov2}, its depth image and its (sparse) instance segmentation mask where only a few objects are segmented. We use the segmentation mask to constrain our motion regularization as well as to obtain a binary foreground/background segmentation, which is used to constrain the motion of the static background (see \cref{subsec:motion_regularizer}).
We provide more details about how to obtain those information in the supplementary material.




\subsection{Monocular Online Tracking Pipeline}
\vspace{-0.1cm}

\label{subsec:method_pipeline}
With our augmented dynamic 3DGS scene representation, we are limited to reconstruct dynamic scenes from a monocular video with known camera poses or that taken from a static camera.
To further enable monocular online point tracking for \textit{pose-free} videos, we build our tracking pipeline on top of SplaTAM~\cite{keetha2024splatam} to simultaneously reconstruct the scene and localize camera poses. 
As shown in \cref{fig:pipeline_overview}, our method takes in a set of required inputs per-timestamp and performs online point tracking with the following steps. 

\PAR{Gaussian and Camera Initialization.}
Given the first frame of a video, we initialize the Gaussians by lifting 2D image pixels to 3D using its pre-computed depth map and its camera pose $\mathcal{C}$ as identity, \ie, the reference of the scene.

\PAR{Camera Optimization.}
We then optimize the camera pose initialized for the current timestamp using the reconstruction losses $L_{rec}$ (\cf \cref{eq:total_rec_loss}) with fixed Gaussian parameters. 
We only apply the loss over pixels that have been observed before and belong to the background which we assume to be static. We identify the prior based on the rendered pixel density which captures the epistemic uncertainty of the reconstruction~\cite{keetha2024splatam} and the latter by the sparse segmentation masks.
As a known limitation to SLAM-based approaches, our method also struggles with extreme camera motions.
When the camera poses are available, we can turn off this step and focus on the Gaussian optimization. 
\PAR{Adding New Gaussians.}
Next, we explore the world by adding a set of new Gaussians using the same initialization defined in the first step. 
We add Gaussians only for sparse and under-represented Gaussian regions identified by a densification mask~\cite{keetha2024splatam}. We provide its detailed definition in the supplementary material.

\noindent\textbf{Gaussian Optimization.} 
Afterwards, we fix camera pose parameters and optimize all 3D Gaussians using a mixture of reconstruction losses defined in \cref{subsec:rec_loss} and 3D regularization losses defined in \cref{subsec:motion_regularizer}, \ie,  $\mathcal{L}_{total} = \mathcal{L}_{rec} + \mathcal{L}_{3D}$.



\noindent\textbf{Gaussian Forward Propagation.} 
After that, we forward propagate the Gaussian means $\mu_i$ and rotations $q_i$ to the next timestamp. 
We assume that close-by Gaussians in the 3D space should move in a similar way if they share similar semantics. Hence, we forward propagate $\mu_i$ based on a kNN-constant velocity assumption:
\begin{equation}
    \mu_i^{\tau+1} = \mu_i^{\tau} + \sum_{j \in \mathcal{N}_i} \sigma(s_{i,j}) (\mu_j^{\tau} - \mu_j^{\tau-1}),
\end{equation}
where $\sigma(s_{i,j})$ is the softmax function over the cosine similarity $s_{i,j}$ between the feature vectors of Gaussian $i$ and a Gaussian $j$ in its k-nearest neighbour Gaussians $\mathcal{N}_{i}$ in 3D space. 
For the rotation $q_i$, we forward propagate it based on a simple constant-rotation assumption:

\begin{equation}
    q_i^{\tau+1} = \Delta q_i^{\tau-1 \rightarrow \tau} * q_i^{\tau} \, ,
\end{equation}
where $\Delta q_i^{\tau-1 \rightarrow \tau} = q_i^{\tau} * (q_i^{\tau-1})^{-1}$ is the relative rotation from time step $\tau-1$ to $\tau$.


\PAR{Camera Forward Propagation.}
Finally, we forward propagate the camera poses for the next timestamp 
by applying the velocity forward projection proposed in~\cite{keetha2024splatam} over the camera pose from the previous timestamp.

For a each frame, we repeat the steps of camera optimization to camera forward propagation.
In the following, we the loss functions to optimize $\mathcal{C}$ and $\mathcal{G}$.


\subsection{Reconstruction Supervision}
\vspace{-0.1cm}

\label{subsec:rec_loss}
As the core supervisory signal for both scene reconstruction and camera localization, we compute reconstruction losses across multiple image types, including RGB color, depth, visual features, and semantics. By leveraging this mixture of information, we impose stronger constraints on point movement over time, requiring consistency with the input frame from multiple semantic perspectives, beyond just photometric errors.
Our reconstruction loss $\mathcal{L}_{rec}$ is specifically defined by a weighted sum over those reconstruction losses:
\begin{equation}
    \mathcal{L}_{rec} = \lambda_I \mathcal{L}_I + \lambda_F \mathcal{L}_F + \lambda_D \mathcal{L}_D + \lambda_B \mathcal{L}_B.
    \label{eq:total_rec_loss}
\end{equation}
We obtain the rendered image of different information by applying the same alpha-composition over different Gaussian attributes $a_i$:
\begin{equation}
    A_p = \sum_{i \in H} T_i \alpha_i a_i,
\end{equation}
with $a_i \in \{c_i, f_i, d_i, g_i\}$ where $d_i$ is the distance of $G_i$ to the camera for rendering the depth map.
%
We use $l2$ distance for feature reconstruction $\mathcal{L}_F$ and $l1$ distance for RGB reconstruction $\mathcal{L}_{I}$, depth reconstruction $\mathcal{L}_{D}$ and background mask reconstruction. 
We ablate the importance of individual reconstruction losses in \cref{subsec:magic_sauce}.

\subsection{3D Regularization}
\vspace{-0.1cm}


\label{subsec:motion_regularizer}
In addition to constraining the optimization in 2D over rendered images, \cite{luiten2023dynamic} introduced three regularization terms applied to a local neighborhood of each Gaussian to enforce the optimization procedure to be physically plausible directly in the 3D space. 
Briefly, they define a short-term local-rigidity loss $\mathcal{L}^{rigid}$ to enforce neighbouring Gaussians to move obeying a similar rigid transform, a local-rotation similarity loss $\mathcal{L}^{rot}$ to force nearby Gaussians to have the same rotations over time, and a long-term local-isometry loss $\mathcal{L}^{iso}$ which constrains the relative distance between two Gaussian means remains the same over time. Due to the space limit, we provide their equations in the supplementary material.


\PAR{Instance-guided Neighbourhood.} \cite{luiten2023dynamic} defines the $k$ nearest neighbours $\mathcal{N}_i$ of a Gaussian $G_i$ based on their 3D distance, which may wrongly define Gaussians from another object to be neighbours around the current object boundary. To avoid this case, we leverage the instance id to select neighbouring Gaussians only within the same object.
%




\PAR{Feature-guided Weighting.}
%
Each regularization loss $\mathcal{L}^{x} \in \{\mathcal{L}^{rigid}, \mathcal{L}^{rot}, \mathcal{L}^{iso}\}$ is further computed by a weighted sum of every Gaussian pair $(G_i, G_j)$, defined by:
\begin{equation}
    \mathcal{L}^{x} = \frac{1}{k|\mathcal{G}|} \sum_{G_i \in \mathcal{G}} \sum_{G_j \in \mathcal{N}_i} w_{i,j} \mathcal{L}_{i,j}^x \, .
\end{equation}
\cite{luiten2023dynamic} defines the pair-wise weight $w_{i, j}=||\mu_i - \mu_j||$ based on their 3D distance. However, different parts of an object can be close to each other but still not connected. Hence they do not move together, \eg, a hand and a leg.
Therefore, we consider to measure the Gaussian distance in the feature space with the cosine similarity between their features $f_i$ and $f_j$, \ie, $w_{i, j}= s_{i,j}$ which significantly improves the tracking performance (\cf \cref{subsec:magic_sauce}).
%

\PAR{Temporal Smoothness Regularization.}
We exploit additional priors that the same Gaussian point should not vary much in color and feature space over time and  background Gaussians should be static. Thus, we restrict individual Gaussian feature $f_i$ and color $c_i$ as well as the mean of background Gaussians to remain close from their values in the previous timestamp using $l_1$ distance. We provide the detailed formula of this smoothness loss $\mathcal{L}_{sm}$ in the supplementary.
%
%

Our total 3D regularization loss is defined as $L_{3D}=\lambda_{sm}\mathcal{L}_{sm} + \lambda_{rigid}\mathcal{L}_{rigid} + \lambda_{iso}\mathcal{L}_{iso} + \lambda_{rot}\mathcal{L}_{rot} $.
We provide ablation study on the importance of all loss functions in \cref{subsec:magic_sauce}.

\subsection{Trajectory Estimation}
\label{subsec:trajectory_est}
\vspace{-0.1cm}






As we perform \textit{tracking-by-reconstruction} with 3D Gaussians, a 2D pixel can be represented via multiple 3D Gaussians. 
Therefore, there exists no one-to-one mapping between the 2D pixel and the 3D Gaussian we track over time. 
To estimate the 2D or 3D trajectory over time for a query pixel $p$, we choose its corresponding  3D Gaussians $G_p$ at time $\tau$ by:

\begin{equation}
    G_p = \min_{G_i \in \mathcal{G}_{v}^{\tau}} (||p - \Pi(W^{\tau} \mu_{i}^{\tau})||_2),
    \label{eq:traj_gaussian_selection}
\end{equation}
where $ \mathcal{G}_{v}^{\tau}$ is the set of Gaussians with visibility $v_i^{\tau} > 0.5$ and $W^{\tau}$ is the camera projection matrix and $\Pi$ is the perspective projection operation. 
%
%
To identify whether a Gaussian is visible, we accumulate the Gaussian opacity in an $\alpha$-composition manner over a set of pixels $P_{G_i}$ that intersect with the 2D Gaussian projection. We thus define the visibility score $v_i^{\tau}$ of a Gaussian as:
\begin{equation}
    v_i^{\tau} = \sum_{p \in P_{G_i}} T_i \alpha_i
\end{equation}
%
Finally, we fully exploit the explicit Gaussian particle representation and obtain the 3D and 2D trajectories by just following $G_p$ through time:
\begin{equation}
    x_p^{3D, \tau} = \mu_p^{\tau}, \quad \quad x_p^{2D, \tau} = W^{\tau} \mu_p^{\tau} 
\end{equation}





\section{Experiments}
\label{sec:experiments}

\begin{figure*}[htp!]
    \centering
    \includegraphics[width=0.95\linewidth]{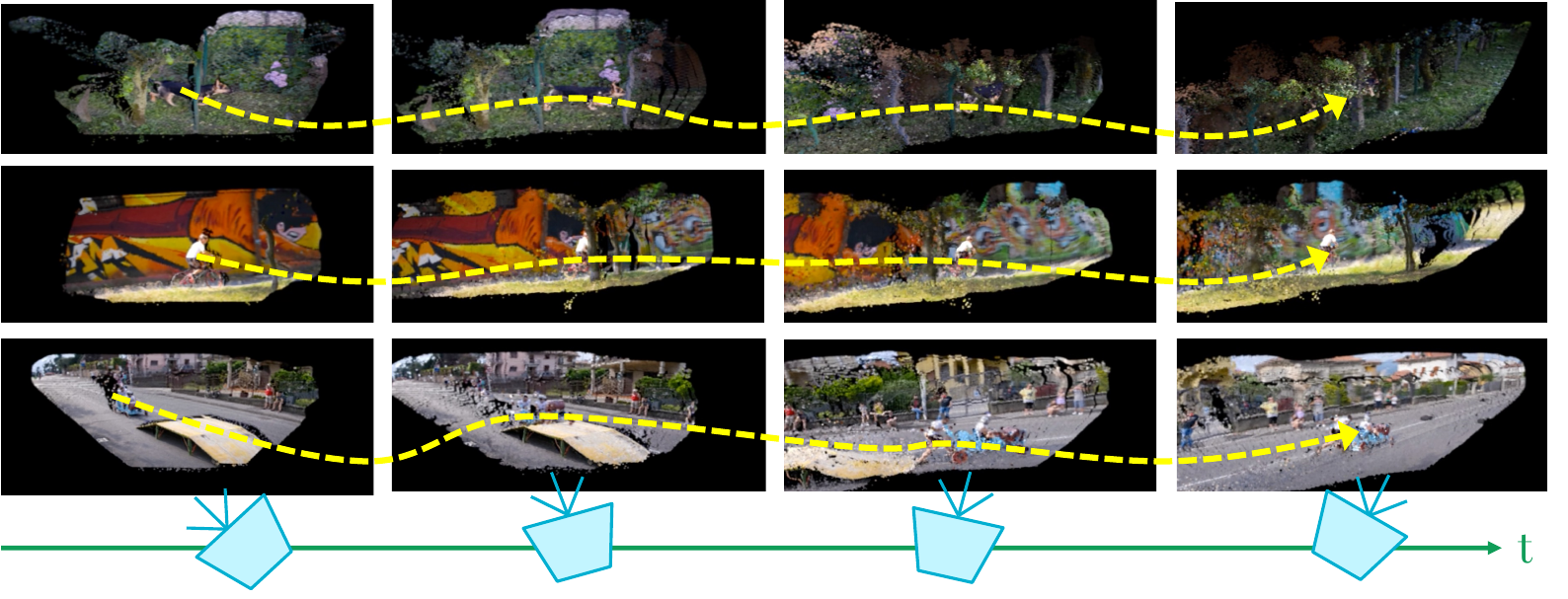}
    \vspace{-0.1cm}
    \caption{\textbf{Increasing the World Online:} Our pipeline is able to gradually add Gaussians to the world. This allows to explore the underlying world as the video progresses. We visualize the increase of the world for two sequences of TAPVid-DAVIS.}
    \label{fig:increse}
\end{figure*}

\begin{table*}[!thp]
    \centering
    \footnotesize
    \setlength{\tabcolsep}{4.2pt}
    \begin{tabular}{l|c|ccc|ccc}
    \toprule
        Method & \# Cameras & MTE$_{2D}$ $\downarrow$ & $S_{2D}$ $\uparrow$ & $\delta_{avg, 2D}$ $\uparrow$ & $MTE_{3D}$ $\downarrow$ & $S_{3D}$ $\uparrow$ & $\delta_{avg, 3D}$ $\uparrow$\\
        \midrule
        \midrule
        D-3DGS \cite{luiten2023dynamic} & 27 & 10.5 & 76.8 & 57.4 & \textbf{13.7} & \textbf{86.8} & \textbf{43.7} \\
        D-3DGS-Mono & 1&  23.3 & 41.0 & 30.6 & 56.0 & 44.5 & 7.3  \\
        \midrule
        \method & 1 & \textbf{6.3} & \textbf{85.7} & \textbf{61.8} & 26.1 & 73.0 & 12.7  \\
        {\method}$^\star$ & 1 & 5.8 & 85.7 & 66.9 & 24.5 & 70.8 & 10.1  \\
    \bottomrule
    \end{tabular}
    \caption{\textbf{Comparison on Panoptic Studio \cite{luiten2023dynamic}:} We compare our performance on PanopticSports as described in \cref{subsec:panopticsports}. D-3DGS-Mono represents the monocular setting of \cite{luiten2023dynamic}. We observe that despite significantly more challenging setting, for 2D point tracking \method outperforms \cite{luiten2023dynamic}. Additionally, \method outperforms D-3DGS-Mono considerably for 3D showing the effectiveness of our method. Finally, {\method}$^\star$ shows that shows that there exists a 3D Gaussian that corresponds even better to the query point.}
    \label{tab:tracking_on_jono}
    \vspace{-0.2cm}
\end{table*}

\begin{table}[t]
    \centering
    \footnotesize
    \resizebox{0.95\linewidth}{!}{
    \begin{tabular}{clcccc}
    \toprule
        & Method & Pretrain & AJ $\uparrow$ &  $\delta_{avg,2D}$ $\uparrow$ & OA $\uparrow$ \\
        \midrule
        \midrule    
       \multirow{6}{*}{\rotatebox{90}{Offline}} 
        & TAP-Net \cite{doersch2022tapvid} & \cmark & 33.0 & 48.6 & 78.8  \\
        & Pips \cite{harley2022particle} & \cmark & 42.2 & 64.8 & 77.7  \\
        & TAPIR \cite{doersch2023tapir} & \cmark & 56.2 & 70.0 & 86.5  \\
        & CoTracker \cite{karaev2023cotracker} & \cmark & 62.2 & 75.7 & 89.3  \\
        & OmniMotion \cite{wang2023omnimotion}  & \xmark & 51.7 & 67.5 & 85.3  \\
        & SpaTracker \cite{Xiao2024spatialtracker} & \cmark & \textbf{61.1} & \textbf{76.3} & \textbf{89.5}  \\
        \midrule        
        \multirow{5}{*}{\rotatebox{90}{Online}} 
        & SpaTracker$^\dagger$ & \cmark & 49.7 & 64.5 & 81.0  \\
        & SplaTAM & \xmark & 13.0 & 22.1 & 72.4 \\        
        & \method & \xmark & 45.8 & 63.1 & 81.1 \\
        & {\method}$^\star$ & \xmark & 52.7 & 69.2 & 84.3 \\
        

    \bottomrule
    \end{tabular}
    }
    \caption{\textbf{Comparison on TAPVID-Davis \cite{doersch2022tapvid} on 2D Point Tracking}: Compared to existing approaches that mostly operate offline and are pre-trained using motion supervision, \method tracks points in an online way without \textit{any} motion signal as supervision. Our emergent motion achieves comparable performance while our oracle baseline ({\method}$^\star$) shows that there exists a 3D Gaussian that corresponds even better to the query point. SpaTracker$^{\dagger}$ represents \cite{Xiao2024spatialtracker} evaluated in an online fashion with time window of 2. 
     } 
   \label{tab:davis}
    \vspace{-0.6cm}

\end{table}

\begin{table}[t]
    \centering
    \footnotesize
    \setlength{\tabcolsep}{4.2pt}
    \begin{tabular}{l|ccc}
    \toprule
        Method  & AJ $\uparrow$ & $\delta_{avg}$ $\uparrow$& OA $\uparrow$ \\
        \midrule
        \midrule
        \method & 45.4 & 63.0  & 81.4 \\
        \midrule
        \midrule
        \multicolumn{4}{l}{\textit{Reconstruction Supervision}}\\

        \midrule
        w/o $\mathcal{L}_B$  & 44.0  & 60.8 & 79.9 \\
        w/o $\mathcal{L}_I$ & 43.2 & 60.4 & 80.0 \\
        w/o $\mathcal{L}_D$ & 37.2  & 52.6 & 71.9\\
        w/o $\mathcal{L}_F$ & 35.9 & 51.1 & 77.2 \\
        \midrule
        \midrule
        \multicolumn{4}{l}{\textit{3D Regularization}}\\
        \midrule
        w/o $\mathcal{L}_{rot}$  & 43.9  & 61.2 & 79.8 \\
        w/o $\mathcal{L}_{iso}$  & 43.7 & 60.9 & 79.1 \\
        w/o $\mathcal{L}_{rigid}$ & 37.9 & 55.3& 77.2 \\        
        w/o instance-guided kNN & 42.3 & 59.6 & 79.5 \\    
        w 3D distance-based weighting  & 39.8 & 55.6 & 75.8 \\
        \midrule
        \midrule
        \multicolumn{4}{l}{\textit{Gaussian Attribute Regularizations}}\\
        \midrule
        w/o kNN forward propagation  & 42.0 & 59.0& 78.6 \\
        w/o temporal smoothness & 42.4 & 59.3& 78.6 \\        
        w/o fixing $s_i$, $o_i$, and $g_i$ & 40.5 & 60.4 & 72.9 \\
    \bottomrule
    \end{tabular}
    \caption{\textbf{Ablation of Importance of Design Choices:} We show the importance of the reconstruction as well as the 3D loss functions. We see that for the prior the embedding as well as the depth losse and for the latter the rigidity loss as well as the semantic weighting are the most important, respectively. We also show the importance of our Gaussian attibute regularizations.
    } 
    \label{tab:ablations}
    \vspace{-0.4cm}
\end{table}


We now evaluate our model for monocular online point tracking on different benchmarks to compare with various baselines. We provide implementation details of our method in the supplementary material.

\subsection{Evaluation on Panoptic Sports}
\label{subsec:panopticsports}
As the first part of our experiment, we compare \method to D-3DGS~\cite{luiten2023dynamic} to highlight the challenge of online monocular point tracking and validate the effectiveness our method. 

\PAR{Dataset.}  To this end, we evaluate our method for both 2D and 3D point tracking on the PanopticSport dataset~\cite{luiten2023dynamic}. It consists of the six sports sequences of the original Panoptic Studio dataset~\cite{joo2015panoptic} where each sequence contains multi-view images taken from 31 temporally aligned cameras with 150 frames being sampled per camera at 30 FPS. 




\PAR{Metrics.} For 2D point tracking evaluation, we report averaged tracking accuracy ($\delta_{avg,2D}$)~\cite{doersch2022tapvid} that computes the mean percentage of points of a trajectory within distance errors of $\{1, 2, 4, 8, 16\}$px to the ground truth trajectory, the median translation error $MTE_{2D}$ in pixels, the survival rate $S_{2D}$ that evaluates the percentage of points below $16$px translation error. For 3D point tracking, we report the median translation errors $MTE_{3D}$ in $cm$, the survival rate $S_{3D}$ at a $50cm$ error threshold and $\delta_{avg, 3D}$ with thresholds of $h = \{1, 2, 4, 8, 16\}$ in $cm$.

\PAR{Baselines.} We compare to the original version of our baseline online tracking approach D-3DGS which is optimized using 27 (out of 31) cameras for each timestamp. 
To reduce to a monocular setting, we choose a single camera such that the tracked object is observed from a reasonable view point. 

\PAR{Results.} As shown in \cref{tab:tracking_on_jono}, D-3DGS-Mono's performance degrades significantly with a $121.9\%$ and $308.8\%$ increase in MTE errors for 2D and 3D point tracking respectively. This highlights that D-3DGS relies heavily on multi-view images to guarantee effectiveness of Gaussian optimization, while \method explores other cues from visual features and instance semantics to constrain optimization signals and compensate the lack of multi-view information.
Comparing both methods between their 2D and 3D tracking performance, we also observe that the monocular setting leads to a bigger drop in performance in 3D compared to D-3DGS optimized in an extreme multi-view setting. 
This reveals that multi-view images provides accurate geometric supervision for 3D reconstruction in general.
In comparison, our method relies on monocular depth maps for geometric supervisions. Thus, we believe more accurate depth maps can largely improve our 3D tracking performance in the future.
Finally, we point out that our method surprisingly outperforms D-3DGS on 2D point tracking in a much more challenging camera setting, which strongly suggests the effectiveness of our design choice. 

\PAR{Influence of trajectory estimation.} 
With our oracle baseline ({\method}$^\star$), we show that there exists a 3D Gaussian that corresponds better to the query point whose trajectory are closer to the ground truth trajectory compared to the one chosen by \cref{eq:traj_gaussian_selection}. 
Thus, our work can be largely improved with a better trajectory estimation mechanism. 
\subsection{Evaluation on TAPVid-DAVIS} 
Different from D-3DGS that requires not only multi-view images but also known camera poses for online tracking, our method is able to model the camera motion jointly with point tracking, which allows us to compare to the state-of-the-art 2D point tracking methods that are evaluated on \textit{pose-free} videos.

\PAR{Dataset.} We evaluate our method on the popular TAPVid-DAVIS dataset introduced as part of the TAPVid benchmark~\cite{doersch2022tapvid}. It consists of 30 highly varying real-world video sequences with unknown camera poses, among which some contains extreme camera motions. This dataset only provides ground truth for 2D point tracking.

\PAR{Metrics.}  We evaluate our 2D point tracking performance utilizing the TAPVid~\cite{doersch2022tapvid} metrics, \ie, averaged tracking accuracy ($\delta_{avg,2D}$) introduced previously, occlusion accuracy (\textbf{OA}) which computes the percentage of frames with correctly predicted visibility and average jaccard (\textbf{AJ}) that combines the both metrics. 

\PAR{Baselines.} While our method focuses on online tracking, we are also curious about the performance gap from the stat-of-the-art offline 2D point trackers. Sepecifically, we compare to Pips~\cite{harley2022particle}, TAP-Net~\cite{doersch2022tapvid}, TAPIR~\cite{doersch2023tapir}, CoTracker~\cite{karaev2023cotracker} and SpaTracker~\cite{Xiao2024spatialtracker} which are all feed-forward methods that require a large amount of data for model pre-training. 
We further consider the recent work OmniMotion~\cite{wang2023omnimotion} that performs test-time optimization over a whole video sequences. Among those method, SpaTracker and OmniMotion are conceptually more similar to ours as they also solve the 2D point tracking by lifting it to 3D, yet with different types of 3D representations.
We also consider a relatively more comparable version of SpaTracker by reducing their sliding window size to 2 (SpaTracker$^\dagger$) for an online point tracking evaluation.
Finally, we compare to SplaTAM~\cite{keetha2024splatam} since we adopted its pipeline and extend it for dynamic scenes.

\PAR{Results.}
As shown in \cref{tab:davis}, when we compare ourselves to offline tracking methods (\textit{upper table}), \method achieves on par tracking accuracy to OmniMotion and outperforms Pips and TAPIR. 
Despite that Omnimotion also performs tracking in 3D, their MLP-based implicit representation requires more expensive optimization over the whole sequence and slow runtime which impedes their application to online tracking ($\sim$18 hrs/sequence on A100 vs \method $\sim$51 min/sequence, \ie, $\sim$45s/frame on RTX 3090). 
While SpaTracker can directly perform online tracking, it leads to $15.5\%$ drop in averaged tracking accuracy since their model is trained to explore cues from multiple temporal frames. 

Within an online tracking category (\textit{lower table}), we significantly outperform our baseline SplaTAM despite using a similar tracking pipeline, since our augmented GS representation is able to handle dynamic objects in the scene.
Compared to SpaTracker, we are slightly lower in performance since their method has been train on a large amount of ground truth data while our method performs online tracking by online 3DGS optimization.  
We show that object motion emerges automatically when we constrain the 3DGS optimization through carefully designed 2D reconstruction supervision and 3D regularization.
Similar to the previous experiment, {\method}$^\dagger$ leads to a boost in performance surpassing SpaTracker, suggesting the need for a better trajectory estimation for evaluation.

\PAR{Qualitative visualization.}
In \cref{fig:increse}, we visualize our gaussian reconstruction across different timestamps. Given our optimized camera pose, we zoom out of the scene and rotate the camera pose slightly.
We show that our method gradually explores the world as the video progress and is able to generate quality 3D reconstruction in complex dynamic scenes from just a single viewpoint frame per timestamp.

\subsection{The Magic Sauce}
\label{subsec:magic_sauce}
To the end of our experiments, we ablate the most important design choices of our approach. 
Please refer to the supplementary material for more ablation experiments.

\PAR{Influence of Reconstruction Losses.}
In \cref{tab:ablations}, we show that our reconstruction losses play a significant role for accurate tracking performance.
Among those, the reconstruction supervision from visual features $\mathcal{L}_F$ and depth maps $\mathcal{L}_D$ are more important.
The rich semantic information from the feature maps enforce the Gaussian motion to agree with its feature observation, while the depth provides signal for geometrically correct reconstruction especially in the absence of multi-view images for monocular online tracking. While RGB information and background masks have lower impact, they still help when the other two signals are noisy due to imperfect off-the-shelf predictions.


\PAR{Influence of 3D Regularization.}
In \cref{tab:ablations}, we further study the importance of our proposed 3D regularization.
Within the three physical-based regularization, the local-rigidity loss $\mathcal{L}_{rigid}$ is more influential than the other two. It forces Gaussians to move rigidly in the 3D space which can prevent 3D Gaussians to move randomly only to fit the 2D observation, which is significant when we lack of multi-view supervisions. 
We also show that our feature-guided regularization weighting is similarly crucial to the rigidity constraint and they cooperate together since Gaussians initialized from monocular depth can be rather noisy. Our feature guidance provides a measure of uncertainty to ignore outlier Gaussian pairs.
Since the visual features are also predicted hence imperfect, instance-guided kNN still improves performance in helping the kNN selection process.


\PAR{Gaussian Attribute Regularization.} Finally, we show that compared to per-point forward propagation, our kNN forward propagation filters outlier motions in previous time steps leading to a significant performance increase. 
The temporal smoothness losses impede the Gaussians from drastically change color and features while enable necessary update for viewpoint changes. 
We also observe that updating $s_i$ and $o_i$ may cause some Gaussians to disappear and thus fix them together with the instance id attribute.



\section{Conclusion}
\label{sec:conclusion}
In this work, we address the task of online point tracking from pose-free monocular camera videos. We introduce \textbf{\method}, a method that achieves online point tracking through dynamic 3D scene reconstruction using 3D Gaussian Splatting alongside simultaneous camera localization. To overcome the challenges of monocular input, we enhance the 3DGS representation with depth maps, instance IDs, and visual features, guiding Gaussian optimization from multiple semantic perspectives and enforcing physically meaningful constraints with robust 3D regularizers. Notably, we demonstrate that accurate point trajectories can emerge from this reconstruction without the need for explicit 2D/3D correspondences as supervision.
Our approach is designed for applications requiring real-time reconstruction and tracking, such as robotic navigation and mixed reality. The ability to extend the 3D Gaussian scene over time as the video progresses enhances online interaction potential. We hope our work inspires the community to explore this path further.
While our method establishes a strong baseline for online point tracking with monocular inputs, it has yet to achieve real-time performance. We anticipate that future work in this area will significantly improve runtime.
\clearpage

{
    \small
    \bibliographystyle{ieeenat_fullname}
    \bibliography{main}
}

\clearpage
\setcounter{page}{1}
\maketitlesupplementary

In this supplementary material, we first provide additional details regarding our method in \cref{sec:method_details} and our implementation in \cref{sec:implementation_details}. 
We further present  2D and 3D point tracking results on the iPhone dataset~\cite{gao2022dynamic} in \cref{sec:iphone} and provide more insights on our design choices with additional ablation studies in \cref{sec:additional_ablations}, followed by discussions and visualizations of our failure cases in \cref{sec:failure_cases}. Finally, we provide additional visualizations of our emergent trajectories and 2D point tracking. We will release our code upon acceptance.

\section{Method Details}
\label{sec:method_details}
\PAR{Densification Mask.} In our approach we exploit the densification mask process from \cite{keetha2024splatam}. This means, we only add new Gaussians for unobserved regions based on a densification mask $M_D(p)$ computed by applying a threshold on pixel densities, \ie, 
\begin{equation}
    M_D(p) = (\sum_{i \in H} T_i \alpha_i) < 0.5,
    \label{eq:densification}
\end{equation}


\PAR{Physically-Based 3D Regularization}
We build on the physically-based priors from \cite{luiten2023dynamic}, \ie, the rigidity $\mathcal{L}_{rigid}$, isometry $\mathcal{L}_{iso}$, and rotation $\mathcal{L}_{rot}$ losses. All losses are applied to every Gaussian directly and computed over its k nearest neighbors (kNN):

\begin{equation}
    \mathcal{L}_{x} = \frac{1}{k|G|} \sum_{i \in |G|}\sum_{j \in kNN_i} w_{i,j} \mathcal{L}_{x,i,j}
\end{equation}
where $x \in \{rigid, rot, iso\}$, $|G|$ is the number of Gaussians, and $w_{i,j}$ is the weight term depending on the Gaussians $i$ and $j$. While \cite{luiten2023dynamic} defines $w_{i,j}$ as the $l_2$ distance in 3D space, we utilize a semantic-based weighting as defined in \textcolor{red}{Sec. 3.3} of the main paper.

The local short-term rigidity loss enforces close-by Gaussians to undergo a similar rigid transformations when observed from Gaussian $i$'s coordinate system. This assumption holds if two Gaussians belong to the same object even for non-rigid objects since it is applied only in a local region. $\mathcal{L}_{rigid}$ is defined by:
\begin{equation}
    \mathcal{L}_{rigid, i, j} = ||(\mu_{j, \tau-1}-\mu_{i, \tau-1}) - \Delta R_i(\mu_{j, \tau}-\mu_{i, \tau})||_2
\end{equation}
 where $\Delta R_i$ is the relative rotation of Gaussian $i$ from time step $\tau$ to $\tau-1$ and is defined as $\Delta R_i = R_{i, \tau-1}R_{i, \tau}^{-1}$. 
 The short-term rotation loss further enforces the rotation of close-by Guassians to undergo similar changes by:

 \begin{equation}
     \mathcal{L}_{rot, i, j} = ||\hat{q}_{j, \tau}\hat{q}_{j, \tau-1}^{-1} - \hat{q}_{i, \tau}\hat{q}_{i, \tau-1}^{-1}||_2
 \end{equation}
where $\hat{q}$ is the normalized quaternion. While this constraint is implicitly enforced by the rigidity loss, we show that additionally adding rotation regularization leads to a slight performance increase (\cf \textcolor{red}{Sec. 4.3} of the main paper).
Finally, the long-term isometry loss additionally enforces Gaussians to keep the initial distance to their kNN Gaussians across time and is hence defined by:

\begin{equation}
    \mathcal{L}_{iso, i, j} = ||\mu_{j, 0}-\mu_{i, 0}||_2 - ||\mu_{j, \tau}-\mu_{i, \tau}||_2 \, .
\end{equation}
%
We compare the impact of individual loss terms in \textcolor{red}{Sec. 4.3} of the main paper.

\PAR{Temporal Smoothness Regularization.} We apply smoothness regularization to Gaussian features $f_i$, Gaussian colors $c_i$ and means of the Gaussains of the background region, defined as:
\begin{equation}
    \mathcal{L}_{sm, f} = \lambda_{f} \sum_{i\in |G|}||f_{i, \tau-1}-f_{i, \tau}||_1 
\end{equation}
\begin{equation}
    \mathcal{L}_{sm, c} = \lambda_{c}\sum_{i\in |G|} ||c_{i, \tau-1}-c_{i, \tau}||_1 
\end{equation}
\begin{equation}
    \mathcal{L}_{sm, \mu_b} = \lambda_{\mu_b}\sum_{i\in |G_b|} ||\mu_{i, \tau-1}-\mu_{i, \tau}||_1
\end{equation}
where $|G|$ is the set of Gaussians, $|G_b|$ is the set of background Gaussians and the final smoothness regularization is given by $\mathcal{L}_{sm} = \mathcal{L}_{sm, f} + \mathcal{L}_{sm, c} + \mathcal{L}_{sm, \mu_b}$.

\section{Implementation Details}
\label{sec:implementation_details}
\PAR{Gaussian Initialization.}
We initialize Gaussian with means $\mu_i$ by lifting the every second pixel defined by its image coordinates to 3D given the depth map and the camera projection matrix. We initialize the rotation to the unit quaternion $q_i = (1, 0, 0, 0)$ and the scale to be dependent on the distance as $s_i = \frac{z}{0.5 * (f_x + f_y)}*(1, 1, 1)$ where $f_x$ and $f_y$ are the focal lengths in $x$ and $y$ direction of the camera. For each pixel position we also unproject RGB color values $c_i \in R^3$, the feature vectors $f_i \in R^{32}$ as well as the instance id $g_i \in R$. This means, we obtain each of the previous information per-pixel and then assign them to its corresponding the unprojected 3D Gaussain.
Since we assume each $G_i$ to be a particle in space, we intialize $o_i$ to be biased towards being visible, \ie, we initialize $o_i=\frac{e^{0.7}}{1+e^{0.7}}$. 

\PAR{Nearest Neighbourhood Selection.} For our Gaussian forward propagation as well as for our semantics guided-weighting of the physically-based priors (see \textcolor{red}{Sec. 3.2} and \textcolor{red}{Sec. 3.4} of the main paper) we require a k nearest neighbor (kNN) set for each Gaussian. We set $k=20$. To determine $\mathcal{N}_i$, we first compute the 2kNN set based on $l_2$ distance in 3D space within all Gaussians belonging to the same instance. We then define $\mathcal{N}_i$ to be those Gaussians from the 2kNN with the smallest $s_{i,j}$, \ie, the smallest cosine distance between the Gaussians' feature vectors. We utilize this kNN set for both, \textcolor{red}{Eq. 3} and \textcolor{red}{7} of the main paper.

\PAR{Gaussian Optimization.} We optimize our dynamic 3DGS model using Adam~\cite{kingma2014adam} optimizer with a learning rate of 0.0016 for $\mu_i$, 0.01 for $q_i$, $0.0005$ for $o_i$, $0.001$ for $s_i$, $0.001$ for $f_i$, $0.0025$ for $c_i$, $0.0001$ for $g_i$, and $0.001$ for camera pose.
We set the reconstruction loss weights to $\lambda_I = 1.0$, $\lambda_F = 16$, $\lambda_D = 0.1$ and $\lambda_{B} = 3$. 
For the physically-based losses, we use the weights to $\lambda_{iso}=16$, $\lambda_{rot}=16$, and $\lambda_{rigid}=128$. 
We set $\lambda_{sm}=1$ and the weighting within the smoothness losses as $\lambda_{f}=20$, $\lambda_{c}=20$, $\lambda_{\mu_b}=5$. 
For experiments on Panoptic Sports we run $500$ iterations per time step while for all other experiments we run $200$ iterations for Gaussian optimization as well as for the camera optimization.

\PAR{Off-the-shelf Data Preparation.}
Given any input RGB image, we extract several other types of information from it using off-the-shelf models. We extract its depth image using the metric depth branch from DepthAnything~\cite{Yang2024depthanything}, as well as visual feature maps using Dinov2~\cite{Oquab2023DINOv2LR} with ViT small backbone followed by PCA to reduce the feature dimension from 384 to 32. For each image, we generate features for five quadratical, overlapping crops to obtain higher resolution feature maps. 
We further assume a sparse instance segmentation masks is provided.
Those data are used either to initialize the scene or provide optimization supervisions.

\PAR{Average jaccard (AJ) metrics.}
\textbf{AJ} combines both metrics by computing the fraction of 
\begin{equation}
    AJ = \sum_{h \in \{1, 2, 4, 8, 16\}}\frac{TP_h}{GT+FP_h}
\end{equation}
where $TP_h$ are the points that lies within the pixel distance threshold h and whose visibility was computed correctly, $FP_h$ are all other predicted points and $GT$ is the number of visible ground truth evaluation points. 

\begin{table*}[!thp]
    \centering
    \footnotesize
    \setlength{\tabcolsep}{4.2pt}
    \begin{tabular}{l|ccc|ccc}
    \toprule
        Method & EPE $\downarrow$ & $\delta_{.05,3D}$ $\uparrow$ & $\delta_{.10,3D}$ $\uparrow$ & AJ $\uparrow$ & $\delta_{avg,2D}$ $\uparrow$ & OA $\uparrow$ \\
        \midrule
        \midrule
        HyperNeRF~\cite{Park2021hypernerf} & 0.182 & 28.4 & 45.8  & 10.1 & 19.3 & 52.0  \\
        DynIBaR~\cite{Li22023dynibar} & 0.252 & 11.4 & 24.6  & 5.4 & 8.7 & 37.7 \\
        Deformable-3D-GS~\cite{yang2023deformable3dgs} & 0.151 & 33.4 & 55.3  & 14.0 & 20.9 & 63.9 \\
        CoTracker~\cite{karaev2023cotracker}+DA~\cite{Yang2024depthanything} & 0.202 & 34.3 & 57.9  & 24.1 & 33.9 & 73.0\\
        TAPIR~\cite{doersch2023tapir}+DA~\cite{Yang2024depthanything} & 0.114 & 38.1 & 63.2  & 27.8 & 41.5 & 67.4 \\
        SOM \cite{som2024shapeofmotion} & \textbf{0.082} & \textbf{43.0} & \textbf{73.3} & 34.4 & 47.0 & \textbf{86.6} \\
        \midrule
        
        \method & 0.161 & 33.5 & 58.1 & \textbf{35.9} & \textbf{58.0} & 65.1 \\
        \method optimized pose & 0.205 & 20.7 & 46.0 & 33.7 & 54.3 & 63.9 \\
        \midrule
        \midrule
        \multicolumn{7}{l}{\textit{Ablating Pose and Depth}}\\
        \midrule
        \method original pose & 0.171 & 32.1 & 55.1 & 35.7 & 56.7 & 66.4 \\
        \method original lidar & 0.198 & 32.9 & 53.0 & 33.2 & 54.5 & 65.0 \\
    \bottomrule
    \end{tabular}
    \caption{\textbf{Iphone Dataset:}. We compare the performance of \method using the aligned DepthAnything \cite{Yang2024depthanything} maps from \cite{som2024shapeofmotion} to other approaches on the Iphone dataset \cite{gao2022dynamic}. Note, prior approaches are all offline and mostly require correspondece-level supervisory signal for motion. 
    We show \method leads to emergent motion in 2D as well as in 3D. Additionally, we show that even with camera pose optimization our EPE is highly competitive compared to the other approaches that all use ground truth or refined camera pose information.}
    \label{tab:tracking_on_iphone}
\end{table*}
\begin{table}[t]
    \centering
    \footnotesize
    \setlength{\tabcolsep}{4.2pt}
    \resizebox{0.95\linewidth}{!}{
    \begin{tabular}{l|ccc}
    \toprule
        Method & AJ $\uparrow$ & $\delta_{avg}$ $\uparrow$ & OA $\uparrow$ \\
        \midrule
        \midrule
        \method & 45.8 & 63.1 & 81.1  \\
        \midrule
        \midrule
        \multicolumn{4}{l}{\textit{Regularization Terms}}\\
        \midrule
        w fixing $c_i$, $f_i$, and $\mu_b$ & 30.8  & 45.3 & 75.4 \\
        w temporal smoothness $o_i$, $s_i$, and $g_i$ & 38.0 & 51.7 & 76.1 \\
        w/o temporal smoothness and fixing & 35.4  & 50.1 & 70.2 \\
        \midrule
        \midrule
        \multicolumn{4}{l}{\textit{Additional Ablations}}\\
        \midrule
        isotropic Gaussians & 42.7  & 59.8 & 79.2 \\
        $\mathcal{L}_{emb}$ w $l_1$ distance & 42.1 & 59.4 & 79.1 \\
        Fixing camera pose  & 40.7 &  57.9 & 77.2 \\
    \bottomrule
    \end{tabular}
    }
    \caption{\textbf{Additional Ablation of Single Part Importance:} In this table, we ablate additional design choices that have less impact on the final performance compared to the ones discussed in \textcolor{red}{Sec. 4.3} of the main paper.
    } 
    \label{tab:ablations_for_supp}
\end{table}
\begin{table*}[!thp]
    \centering
    \footnotesize
    \setlength{\tabcolsep}{4.2pt}
    \begin{tabular}{l|ccc|ccc|ccc}
    \toprule
        Method & $MTE_{2D}$ $\downarrow$ & $S_{2D}$ $\uparrow$ & $\delta_{avg, 2D}$ $\uparrow$ & 2D \%1 & 2D \%8 & 2D \%16 & $MTE_{3D}$ $\downarrow$ & $S_{3D}$ $\uparrow$ & $\delta_{avg, 3D}$ $\uparrow$\\
        \midrule
        \method-DA \cite{Yang2024depthanything} & \textbf{3.7} & 83.7 & 59.0 & 19.6 & 80.3 & 88.7 & 70.7 & 25.4 & 0.7\\
        \method-D-3DGS-D & 6.3 & \textbf{85.7} & \textbf{61.8} & \textbf{25.3} & \textbf{83.6} & \textbf{90.3} & 26.1 & 73.0 & 12.7  \\
    \bottomrule
    \end{tabular}
    \caption{\textbf{Impact of Different Depth Predictions:} We compare utilizing DepthAnyhthing \cite{Yang2024depthanything} metric depth prediction with rendered depth predictions from \cite{luiten2023dynamic} for the Panoptic Sports dataset \cite{luiten2023dynamic}. Additionally to the main metrics, we also report the percentage of points within $1, 8$ and $16$px distance. We observe that for 2D point tracking the depth prediction approach does not have a major impact on the overall performance. Meanwhile the performance drop is more significant for high precision metrics, \eg, $2D 1\%$. Additionally, for 3D point tracking, we observe a severe performance drop with respect to all metrics.}
    \label{tab:jonodepthpred}
\end{table*}

\begin{table*}[t]
    \centering
    \footnotesize
    \setlength{\tabcolsep}{4.2pt}
    \begin{tabular}{l|ccc|ccc}
    \toprule
        Method & $MTE_{2D}$ $\downarrow$ & $Survival_{2D}$ $\uparrow$ & $\delta_{avg, 2D}$ $\uparrow$ & $MTE_{3D}$ $\downarrow$ & $Survival_{3D}$ $\uparrow$ & $\delta_{avg, 3D}$ $\uparrow$  \\
        \midrule
        \midrule
        $\alpha$-composition \cite{som2024shapeofmotion} & 7.2 & 81.9 & 55.0 & 32.1 & 65.3 & 8.7  \\
        Closest 3D Gaussian & 9.4 & 79.6 & 35.7 & 26.1 & 73.0 & 12.7 \\
        \method (Closest 2D Projection) & 6.3 & 85.7 & 61.8 & 24.1 & 71.0 & 10.1 \\
    \bottomrule
    \end{tabular}
    \caption{\textbf{Choice of Gaussians in Trajectory Estimation:} We compare different approaches for choosing the Gaussian to track. We find that for 2D point tracking, our proposed choice of Gaussians based on the closest 2D Guassian projection achieves the best performance. The performance of $\alpha$-compositioning lags behind since it chooses Gaussians per timestamp without enforcing those Gaussians to belong to the same point in 3D. 
    When metric depth measurements are available, we choose the closest 3D Guassian directly in 3D, leading to improved 3D tracking performance.
    }
    \label{tab:gauss_choice}
\end{table*}

\section{Comparison on iPhone Dataset}
\label{sec:iphone}
To further comprehensively understand \method's performance, we evaluate it for 2D and 3D point tracking on a third benchmark, the iPhone dataset \cite{gao2022dynamic} in \cref{tab:tracking_on_iphone}.

\PAR{Dataset.} 
The iPhone dataset \cite{gao2022dynamic} contains moving camera, casual captures of real-world scenes recorded using an iPhone. This setting is highly aligned with potential use cases of mixed reality. Compared to Panoptic Sports \cite{luiten2023dynamic} and TAPVID-Davis \cite{doersch2022tapvid}, the dataset provides rgb images, lidar depth, camera poses of the moving camera, as well as sparse 2D point correspondences across the entire video to evaluate 2D and 3D point tracking.
%

\PAR{Metrics.} We follow the same evaluation protocol as in \cite{som2024shapeofmotion} where the sparse correspondences are forward and backward tracked across the sequence. 2D point tracking is evaluated using the TAPVID-Davis metrics, \ie, AJ, $\delta_{avg, 2D}$ as well as OA. However, instead of using thresholds of $\{1, 2, 4, 8, 16\}$px, the authors use $\{4, 8, 16, 32, 64\}$px. For 3D point tracking, the authors evaluate the end point error (EPE) which is the average $l_2$ error in 3D reported in $m$ over predicted trajectories. Additionally, they report the fraction of points within $5cm$ and $10cm$ to the ground truth trajectories, \ie, $\delta_{.05, 3D}$ and $\delta_{.10, 3D}$. 

\PAR{Baselines.}
Following SOM~\cite{som2024shapeofmotion}, we use their aligned DepthAnything \cite{Yang2024depthanything} depth maps. We downscale input images by a factor of $0.5$, initialize Gaussians for every pixel position and adapt the visibility threshold accordingly to $0.1$.
We compare two variants of our method. For one variant, we use the refined camera poses from SOM and for the other one, we use our optimized camera poses.
As shown in \cref{tab:tracking_on_iphone}, we compare ourselves to a set of existing \textit{offline} 2D/3D point trackers. 
Being highly relevant to our approach, SOM~\cite{som2024shapeofmotion} also leverages 3D Gaussians as a dynamic scene representation and
could be viewed as an offline version of \method.
However, the authors optimize their approach utilizing point trajectories extracted by TAPIR \cite{doersch2023tapir} as supervision signal and, hence, use correspondence-level supervisory signals similar to CoTracker, TAPIR. 
In contrast to them, our online tracker \method shows emergent motion from online 3D Gaussian reconstruction.

\PAR{Refined Poses from \cite{som2024shapeofmotion}.} Despite the online character of our method and not requiring correspondence-level supervision, \method achieves on-par performance with many existing approaches in 2D as well as in 3D. Interestingly, for $\delta_{avg, 2D}$ and AJ we even outperform all previous methods. In 3D, we perform better than the purely view-reconstruction approaches, \ie, those approaches that do not require correspondence-level supervision. Additionally, we perform on par with CoTracker\cite{karaev2023cotracker}+DA\cite{Yang2024depthanything} underlining \method's ability to generate emergent trajectories.

\PAR{Optimized Poses.} Even if we optimize for camera pose additionally, our EPE is highly competitive despite all other approaches despite them utilizing ground truth or refined camera poses. However, due to inaccuracies in our optimized camera poses we observe a drop in performance especially for the high precision metrics like $\delta_{.05,3D}$. Evaluating our accuracy of our optimized camera poses using ATE RMSE, \ie, average translation error measures in root mean square error we obtain an ATE RMSE of $10.49 cm$  mirroring the difficulty of optimizing camera poses for dynamic sequences.

\PAR{Ablating Aligned Depth and Refined Pose \cite{som2024shapeofmotion}.} We show that utilizing the original LiDAR singal to supervise \method and the original poses provided by the iPhone dataset leads only to a slight performance decrease in 3D as well as in 2D mirroring \method's robustness. 

\section{Additional Ablation Studies}
\label{sec:additional_ablations}
In this section, we provide additional ablation studies in \cref{tab:ablations_for_supp} to understand our method more comprehensively.

\PAR{Temporal Smoothness Regularization Terms.} 
We provide additional ablation studies on the temporal smoothness regularization. To recap, in our setting we apply temporal smoothness on $f_i$, $c_i$ and $\mu_{b}$ while we fix $s_i$, $o_i$ and $g_i$ over time. In the main paper we show the performance drop of not applying temporal smoothness, \ie, optimize $f_i$, $c_i$ and $\mu_b$ at every time step without additional supervision.We also show the performance drop of not, fixing $s_i$, $o_i$ and $g_i$, \ie, also optimizing them at every time step. 
We now also provide experiments for fixing  $f_i$, $c_i$ and $\mu_{b}$, applying temporal smoothness on $s_i$, $o_i$ and $g_i$ as well as not applying temporal smoothness nor fixing values at all (see section \textit{Regularization Terms} in \cref{tab:ablations_for_supp}). We observe that all three lead to significant performance degradation: (i) fixing $c_i$, $f_i$, and $\mu_b$ leads to \method not being able to adapt to, \eg, temporal inconsistencies in image feature prediction or slight color changes due to viewpoint changes; (ii) temporal smoothness terms on $o_i$, $s_i$, and $g_i$ can lead to, \eg, Gaussians disappearing due to scale or opacity changes; (iii) not applying any regularization allows the Gaussians to change their attributes freely which allows them to "cheat" to adapt to the a new time frames supervisory signal.

\PAR{Isotropic vs. Anisotropic Gaussians.}
Instead of using anisotropic Gaussian distributions, \cite{keetha2024splatam} chose to use isptropic Gaussian distributions for their use case of generating maps of static scenes. However, for non-rigid objects, this choice is less suited for two reasons: (i) composing non-rigid objects and motion with isotropic Gaussians, \ie, small spheres, restricts the degrees of freedom to adapt the geometry and geometrical changes; (ii) the rotation of Gaussians actually does not matter for the reconstruction losses, hence, it adds noisy signals in supervision signal. This negatively impacts the rigidity loss, for which the rotation signal is of major importance.

\PAR{Penalizing Outliers for the Feature Map Reconstruction.} 
Our experiments show that utilizing rendered feature maps leads to significant performance improvement. Rendering feature maps can be seen as a stronger, less ambiguous supervisory signal than RGB colors that pulls the Gaussians to the correct location to match a given time steps observation. We found that utilizing a $l_2$ distance in computing the feature reconstruction loss 
penalizes outlier Gaussians in a stronger way (see "$\mathcal{L}_{emb}$ w $l_1$ distance" in \cref{tab:ablations_for_supp}).

\PAR{Disentangled Camera Motion.} 
\cite{wang2023omnimotion} showed that it is not necessary to explicitly model camera motion, but it is sufficient to entangle camera and object motion. However, in \cref{tab:ablations_for_supp} we demonstrate that not explicitly modeling camera motion, \ie, fixing camera pose, leads to a significant performance drop for \method. Additionally, this feature is important for real-world applications that require interaction with the real world. 


\PAR{Different Depth Priors}
In our online monocular setting, the depth information is directly used for the Gaussian mean initialization as well as for the depth reconstruction loss meanwhile also influences the physics-based loss functions. 
%
Therefore, we study the impact of the depth map on our tracking performance by comparing our \method using the zero-shot metric depth predictions from DepthAnything \cite{Yang2024depthanything} (DA) with using the depth predictions from \cite{luiten2023dynamic} (D-3DGS-D).
Additionally to the metrics from the main paper, we also report the percentage of points within $1, 8$ and $16$px distance. 
As shown in \cref{tab:jonodepthpred}, we only observe slight performance difference using different depth maps for 2D point tracking, since a correct 2D trajectory does not necessarily require a correct 3D trajectory. This performance difference is more observable in high precision metrics, \ie, fraction of points within $1$px distance.
In contrast, for 3D point tracking, we observe a significant performance increase with better depth estimation, \ie, D-3DGS-D.
This suggests that our method directly benefits from future monocular depth estimation advancement.

\PAR{Choice of Gaussians in Trajectory Estimation.}
We validate our way of choosing Gaussians for trajectory estimation (as introduced \textcolor{red}{Sec 3.5} of the main paper) by comparing to other ways of choosing Gaussians as introduced by other works.
Inspired by \cite{wang2023omnimotion},  SOM\cite{som2024shapeofmotion} computes the 3D trajectory of a corresponding pixel $p$ at time $\tau$ by taking the set of Gaussians $H(p)$ into account that intersect with pixel $p$ at $\tau_s$:

\begin{equation}
    X_{p}^{\tau_s \rightarrow \tau} = \sum_{i \in H(p)} T_i \alpha_i \mu_{i}^{\tau}
\end{equation}
where $\mu_{i}^{\tau}$ is the mean of $G_i$ at time $\tau$ and $X_{p}^{\tau_s \rightarrow \tau}$ is the 3D location at time $\tau$ corresponding to the trajectory starting from $p$ at $\tau_s$. For 2D point tracking, the authors project $X_{p}^{\tau_s \rightarrow \tau}$ to the image plane:

\begin{equation}
    X_{p, 2D}^{\tau_s \rightarrow \tau} = W^{\tau} X_{p}^{\tau_s \rightarrow \tau}
\end{equation}
where $W_{\tau_s}$ is the viewing transformation at $\tau$. 
We compare \method to the above explained approach from \cite{som2024shapeofmotion} denoted as $\alpha$-composition. 
Additionally, assuming 3D ground-truth (GT) trajectories are available, instead of choosing a 3D Gaussian for a query pixel based on its closest 2D projection, we directly choose the 3D Gaussian closest to the GT 3D Gaussian for tracking.
%
As shown in \cref{tab:gauss_choice}, $\alpha$-composition leads to worse performance in general, as the 3D Gaussians are selected per timestamp and thus can belong to different points in 3D. 
While choosing the Gaussian based on the 3D means performs slightly better for evaluation in 3D tracking, it's performance in 2D is worse. 
Furthermore, 3D GT trajectory information is not generally available, \eg, for TAPVID-Davis.
We therefore stick to our proposed choice of Gaussian as in  \textcolor{red}{Sec 3.5}. 

\begin{figure*}
    \centering
    \includegraphics[width=0.95\linewidth]{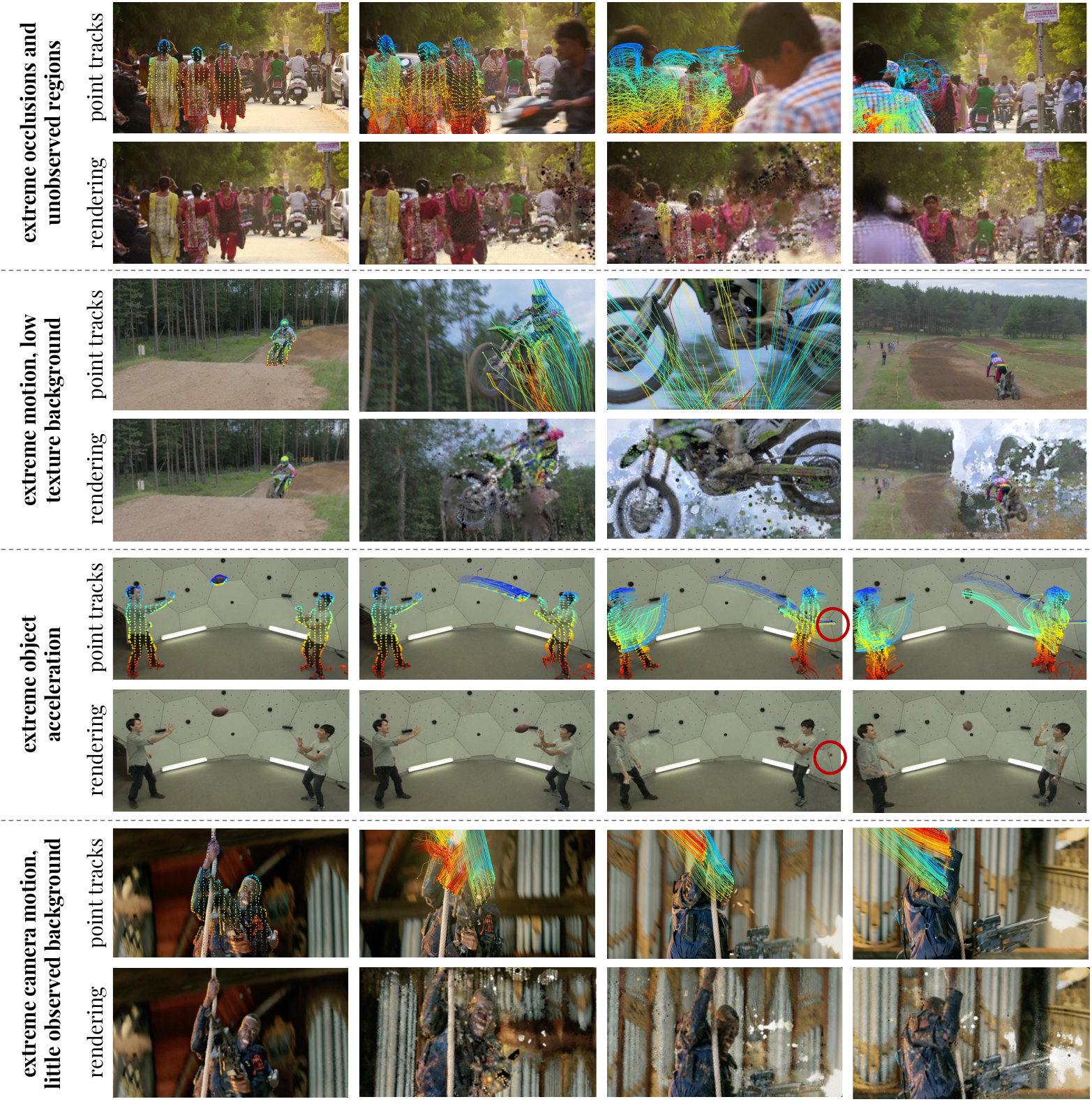}
    \vspace{-0.3cm}
    \caption{\textbf{Failure Cases of \method:} \method struggles (i) to track points and add new Gaussians in sequences with extreme occlusions; (ii) to track the camera position as well as the Gaussians in sequences with extreme camera and object motion as well as low background texture; (iii) to track points when extreme acceleration changes occur; (iv) to track Gaussians and camera positions when solely little background is observed but extreme camera motion occurs.}
    \vspace{-0.3cm}
    \label{fig:failure_cases}
\end{figure*}

\begin{figure*}
    \centering
    \includegraphics[width=0.85\linewidth]{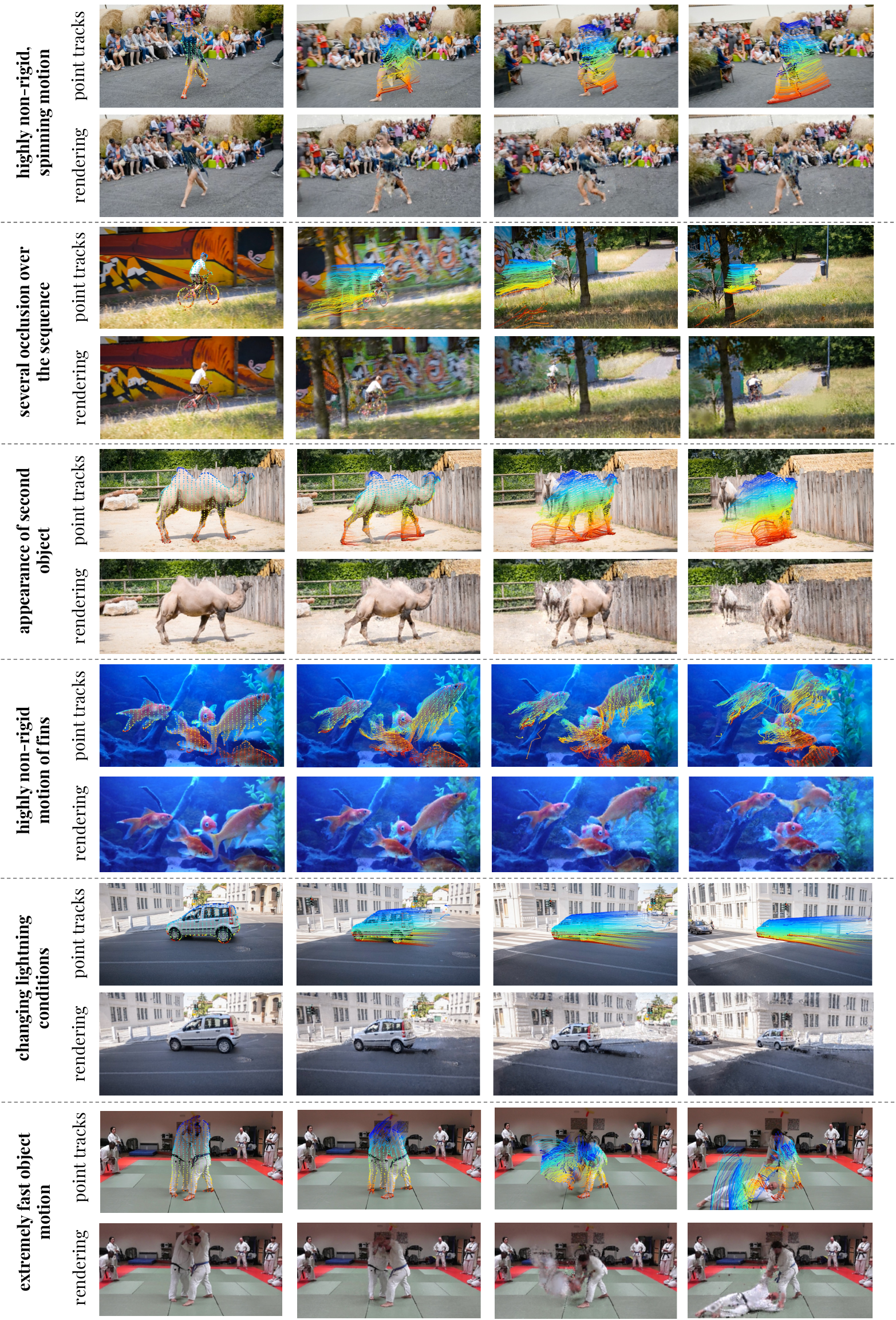}
    \caption{\textbf{Visualizations on TAPVID-Davis:} We visualize renderings as well as point tracks on challenging scenes on TAPVID-Davis. \method is able to generate emerging trajectories despite facing non-rigid motion, occlusions, appearance of new objects, or fast motion.}
    \label{fig:visualizations}
\end{figure*}

\begin{figure*}
    \centering
    \includegraphics[width=0.85\linewidth]{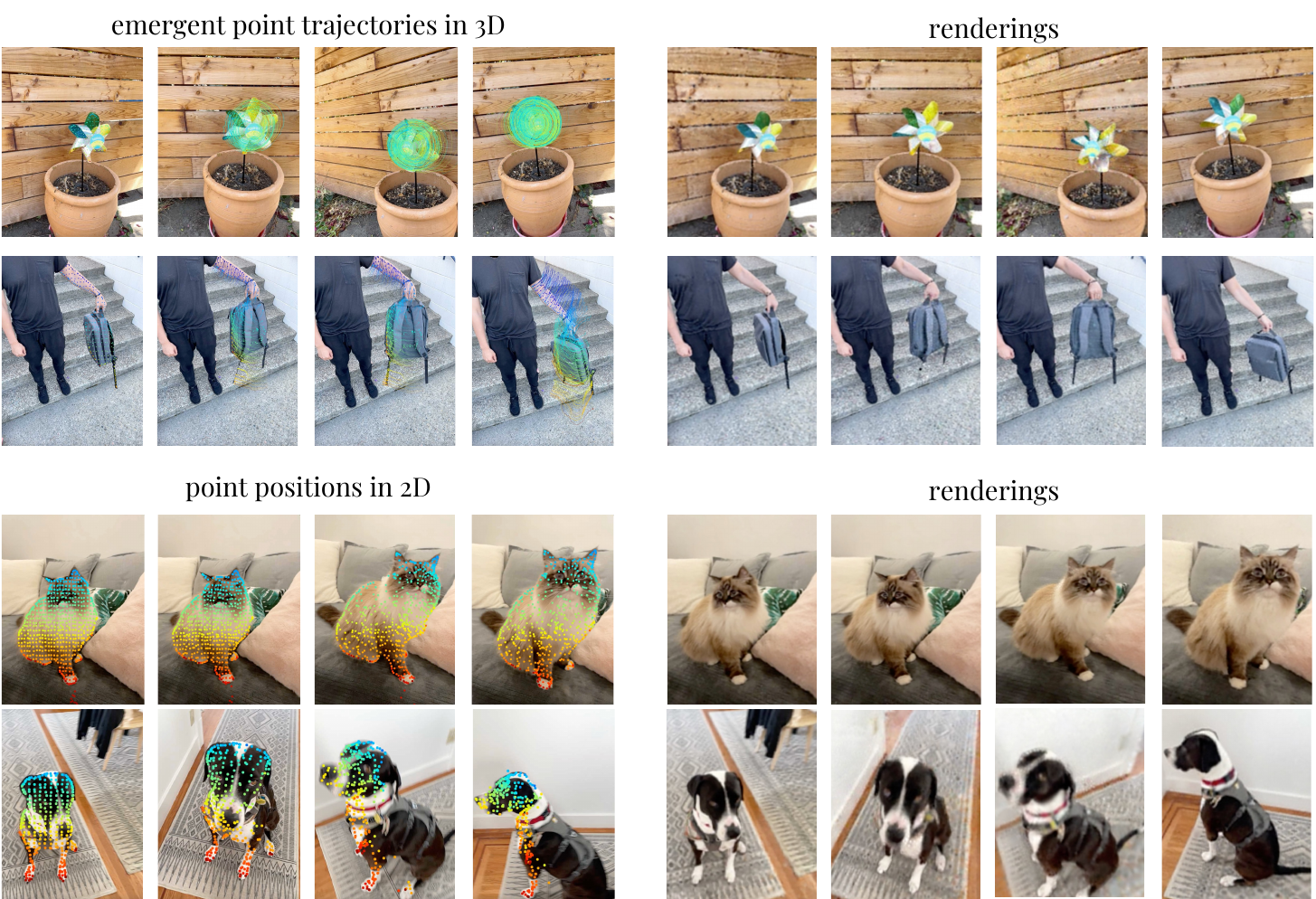}
    \caption{\textbf{Visualizations on iPhone Dataset:} We visualize renderings as well as point trajectories and point positions on casual captures from the iPhone dataset. \method is able to generate trajectories for challenging spinning motions and is able to track points in 2D.}
    \label{fig:visualizations_iphone}
    \vspace{-0.3cm}
    
\end{figure*}

\section{Failure Cases}
\label{sec:failure_cases}
\vspace{-0.1cm}
In this section, we discuss and visualize several failure cases of our method in \cref{fig:failure_cases} as well as point out potential solutions. We hope those insights spark future research along this direction to address those limitations.

\PAR{Extreme Camera Motion.} If camera poses are unknown, we estimate the camera poses. However, for extreme camera motion as well as background with little texture, \method struggles to reconstruct the motion appropriately which is also a common failure case in SLAM methods. 

\PAR{Extreme Occlusions.} In cases with sudden extreme occlusions, \method struggles to reconstruct the camera pose. Additionally, due to our online monocular character, \method struggles to track points over long-term occlusions. Utilizing a stronger constant velocity assumption or exploiting fine-grained instance id's can potentially help to recover after occlusions. 

\PAR{Extreme Object Acceleration.} For extreme object acceleration, \method struggles to model the sudden change of motion. This is due to the fact that we forward propagate points to the next frame and our constant velocity assumption breaks for sudden extreme velocity changes. Future research could explore utilizing motion prediction models for forward propagation to resolve this issue.

\PAR{Previously Observed Regions.} When adding new Gaussians based on the densification concept, it can happen that a new object enters the scene in front of a previously observed concept. In such cases, no new Gaussians will be added for the newly appearing concept. Towards this end, \cite{keetha2024splatam} also exploits depth errors to add Gaussians. However, this requires highly accurate depth prediction, which we do not have access to in, \eg, the TAPVIS-Davis sequences. We assume the advances in depth prediction models will help in resolving this challenge.

\PAR{Using Depth As Supervisory Signal.} Since we are operating in a monocular and online setting, depth predictions are the most important 3D information for \method. However, noisy depth maps can lead to degrading performance when using them directly as supervisory signal, \ie, using a rendered depth loss. Hence, we only exploit it moderately for supervision. Additionally, we add Gaussians by lifting pixel positions to the 3D space given using depth maps. However, since those depth maps are not precisely consistent over time, the Gaussians may be added in a highly incorrect location in the 3D space from which it is highly challenging to recover. 
As also suggested by our ablation in \cref{tab:jonodepthpred}, the recent advances in depth prediction models will support in solving this issue. 

\PAR{Spinning and Turning Objects.}
Similarly to adding Gaussians for newly appearing objects, adding Gaussians for a newly appearing side spinning objects is highly challenging since previously existing yet now occluded Gaussians can be re-used to "cheat". Exploiting recent approaches in zero-shot mesh prediction could help to populate the whole mesh, \ie, the whole surface of a spinning object with Gaussians.

\section{Additional Visualizations}
Finally, we present additional visualizations on TAPVID-Davis in \cref{fig:visualizations} as well as on the iPhone dataset \cref{fig:visualizations_iphone}. For the TAPVID-Davis, we visualize emergent trajectories for scenes with occlusions, appearing objects, highly non-rigid motions, changing lightning conditions as well as extremely fast object motion. On the iPhone dataset we show that \method is able to generate emergent trajectories as well as to track points in a robust manner in 2D.

\end{document}